# Controlling Grokking with Nonlinearity and Data Symmetry

Ahmed Salah[*], David Yevick

*Department of Physics, University of Waterloo, ON N2L 3G1, Canada*

*Corresponding author email address: asalah@uwaterloo.ca

*Abstract–* This paper demonstrates that grokking behavior in modular arithmetic with a modulus *P* in a neural network can be controlled by modifying the profile of the activation function as well as the depth and width of the model. Plotting the even PCA projections of the weights of the last NN layer against their odd projections further yields patterns which become significantly more uniform when the nonlinearity is increased by incrementing the number of layers. These patterns can be employed to factor *P* when *P* is nonprime. Finally, a metric for the generalization ability of the network is inferred from the entropy of the layer weights while the degree of nonlinearity is related to correlations between the local entropy of the weights of the neurons in the final layer.

## 1. Introduction

The phenomenon of "grokking" during neural network training was first observed and empirically described in the context of Transformers by Power et al. (2022) [1]. Before the onset of grokking, the model exhibits a high level of training accuracy but a low test accuracy. After further training, however, the test performance improves markedly. Similar behavior is observed in double descent which is also associated with the difference in the time required to learn dissimilar patterns in the training test sets. Davies et al. (2023) [2].

In the context of simple physical or algorithmic models, grokking provides a controlled environment in which to examine the conditions required to learn underlying relations in the dataset. These include, but are not limited to, the similarity in the table of the dataset, the data set size, training fraction and the influence of nonlinearity. As well, the physical meaning of collective properties such as the PCA projections of neural network (NN) weights can be investigated as well as the correlations between the weights of the neurons.

Grokking has recently been studied extensively Davies et al.[2], 2023, Nanda et al., 2023[3], Thilak et al.[4], 2022, Varma et al.[5], 2023, Liu et al., 2022a[6],b[7], Gromov, 2023[8], Schaeffer et al., 2023[9], Michaud et al. 2023[10]) and several semi-analytical frameworks have been advanced. While some of these require weight decay [Liu et al., 2022[6], Varma et al., 2023[5]], a two-layer MLP without weight decay also exhibits grokking Kumar et al. [11]. In [4], grokking without regularization was accompanied by a "slingshot" event—a sudden spike in training loss followed by improved generalization. The analysis of Varma et al. [5] and Liu et al. [7] do not incorporate slingshot or oscillation behavior but instead is based on the relative strength of memorization and generalization. Liu et al. [7] attribute generalization to learning a representation of the input embeddings which, in the case of algorithmic datasets, is generally accompanied by the emergence of structure in the embeddings. This manifests itself as a circular pattern when consecutive PCA components of the inputs are plotted against each other. In this paper, however, a similar generalization behavior is achieved in a standard MLP in the absence of embeddings and transformers while it is further observed that grokking can occur without inducing an observable structure in PCA space.

The principal conclusions of this paper are the following:

- Grokking occurs even if the neurons properties are aperiodic.
- The interaction among the neurons within a layer is quantified by a correlation function that changes its properties.
- The nonlinearity of the NN, which depends on the form of the activation function and the width and depth of the network, can be employed to control the grokking behavior.
- The NN nonlinearity further impacts the symmetries inherent in the PCA eigenvectors of the NN weights.
- These symmetries, especially for small datasets, can yield a factorization of the modulus.
- Grokking also occurs when the PCA projections are not manifestly symmetric.
- Grokking and the effect of nonlinearity can be partially characterized by the entropy of the NN weights as well as their mutual correlations.

## 2. Preliminaries and Setup

**Grokking modular arithmetic**. The calculations of this paper apply a two-layer MLP without embedding to data generated through modular addition, $Y = (i + j)\%P$, for $i, j = 0, 1, \ldots P-1$ where $i$ and $j$ are separately one-hot encoded. These vectors are then concatenated to generate a single tensor $X$ of length $2P$. The dimension of the output vector is $P$, while the size of the dataset is $P^2$. Since this corresponds to a classification problems with the label $Y$, cross entropy loss is utilized together with an Adam optimizer (Loshchilov & Hutter, 2019) [12] with weight decay, as weight decay has been proven to significantly impact grokking (Power et al., 2022[1]; Nanda et al., 2023[3]; Varma et al., 2023[5]).

For a quadratic activation function the network mapping becomes (Gromov, 2023) [8],

$$f(x) = \frac{1}{DN} W^{(2)} (W^{(1)} x)^2 \tag{1}$$

where $D$ is the input dimension, $N$ is the hidden layer width, and $W^{(i)}$ are the $i^{th}$ layer weights matrices. The model size is varied by adjusting the hidden size dimension, HIDDEN_DIM, while the fraction of the data employed for training data is TRAIN_FRAC. The MLP utilized here consists of two layers; the input layer of size (HIDDEN_DIM) and the output one of size ($P$). The effect of inserting an intermediate layer of size (HIDDEN_DIM) will also be considered.

The weights $W^{(1)}_{kn}$ and $W^{(2)}_{qk}$ that are consistent with modular addition problem are given by

$$W^{(1)}_{kn} = \begin{pmatrix} \cos\left(2\pi \frac{k}{p} n_1 + \varphi^{(1)}_k\right) \\ \cos\left(2\pi \frac{k}{p} n_2 + \varphi^{(2)}_k\right) \end{pmatrix}^T, n = (n_1, n_2) \tag{2}$$

$$W^{(2)}_{qk} = \cos\left(-2\pi \frac{k}{p} q - \varphi^{(3)}_k\right) \tag{3}$$

with $n_1, n_2 = 0, 1, \ldots p-1$. For inputs $(n, m)$, the unnormalized outputs of the first layer are

$$h^{(1)}_k(n, m) = \cos\left(2\pi \frac{k}{p} n + \varphi^{(1)}_k\right) + \cos\left(2\pi \frac{k}{p} m + \varphi^{(2)}_k\right) \tag{4}$$

where for modular addition

$$\varphi_k^{(3)} = \varphi_k^{(1)} + \varphi_k^{(2)} \tag{5}$$

This ensures that the terms of $h_q^{(2)}(n,m)$, interfere constructively and are therefore described by

$$\boldsymbol{h}_q^{(2)}(n,m) \approx \frac{1}{2} \sum_{k=1}^{N} \cos\left(2\pi \frac{k}{p}(n+m-q)\right) \tag{6}$$

Additionally, degree 2 polynomial activation functions of the form $\emptyset(x) = bx + ax^2$, which have been shown to provide additional accuracy when suitably optimized (Yevick, 2024) [13] will be considered.

Barak et al. (2022) [14] introduced and theoretically analyzed metrics that are associated with emergent behavior. Here a metric is introduced to quantify the extent of grokking of a network based on of the weights of the last layer. This differs from the progress measures suggested by Nanda [3] that are based on mechanistic interpretability but is consistent with results for the entropy associated with the explained variance ratio derived from the PCA of the embeddings obtained by Liu et al. (2022) [7].

## 3. Experiments and Findings

### 3.1 Weights for Different Connections of the NN and Correlation of Neurons

The harmonic behavior of the weights described in the previous section was observed in many of our numerical calculations. In particular, the weights of the connections from each neuron in the last layer to a single neuron in the first layer as illustrated in Fig. 1 is quasi-sinusoidal for some of the first layer neurons while the corresponding Fourier spectrum in Fig. 2 is sharply peaked at the positive and negative frequencies present in the cosine function. However, the weights for the connections to other first layer neurons can be nonperiodic without eliminating the grokking behavior as seen for instance with P=27 in Fig.2, in apparent contradiction to (Gromov, 2023) [8]. If weight decay is employed, a certain fraction of the last layer connection weights decreases below $10^{-41}$ many epochs after grokking as shown in Fig.3, without affecting the test accuracy. Evidently, the surviving connections are those that convey the largest fraction of the information present in the data.

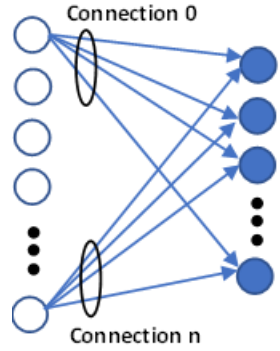

Figure 1 Groups of neural network connections whose weights are analyzed below

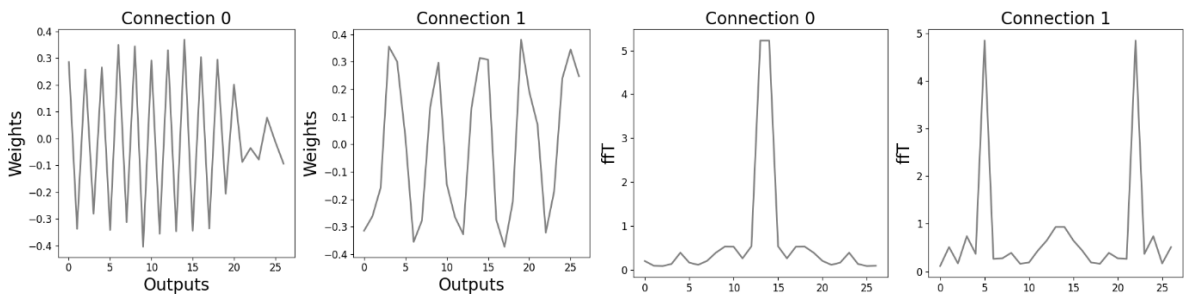

Figure 2 The weights of the output neurons (a) and the associated Fourier transform (b) where each figure represents the first two connections of neurons with the first layer.

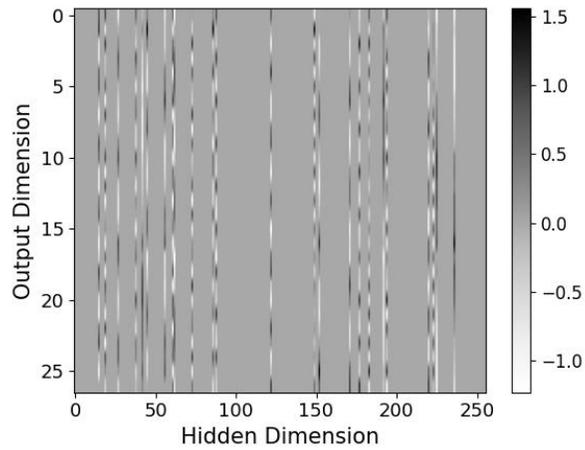

Figure 3 The weights of the last layer in the presence of weight decay after many epochs.

The degree of periodicity in the model weights can be quantified by evaluating correlation function of the weights given by

$$Corr(k) = \sum_{l=0}^{N-1} x(l)x(l-k) \tag{7}$$

where $x(l)$ is the value of the input sequence, weights of the connections of the last layer nodes, at index $l$, while $x(l-k)$ is the value of the sequence shifted by $k$ in the presence of periodic boundary conditions. The length of the sequence is $N$ and the correlation at each point of overlap between the sequences is calculated and the values are then normalized. The resulting correlation of the weights of the neurons in the output layer that are associated with the connections to a single neuron in the first layer is described for certain output notes by the damped oscillating function of Fig. 4. However, other sets of connections to the first layer do not display a harmonic behavior. Since a positive correlation indicates heuristically that the network operates in a similar fashion when applied to certain input features, the four oscillations in Fig. 4 are plausibly related to the number of rotations in the PCA diagrams. The rate of decay of the correlation could then be associated with the deviation of the rotational patterns from perfectly repeating structures. The groups of connections with oscillating correlations would then be associated with the nodes that have learned the periodicity in the data resulting from the modular addition.

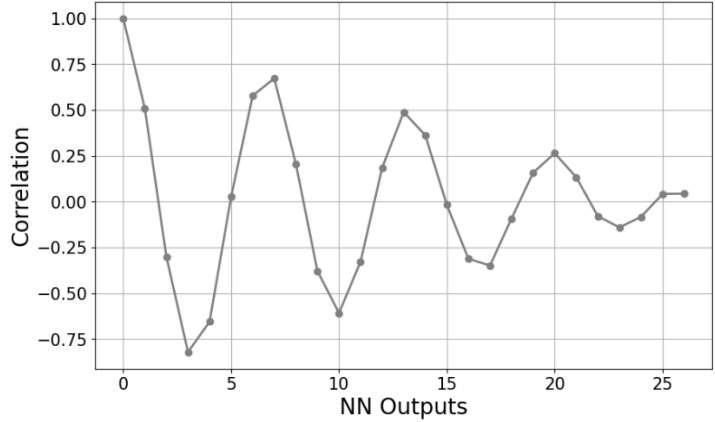

*Figure 4 The autocorrelation of the weights of the neural network connections to one of the nodes of the last layer.*

### 3.2 Activation Function

Changing the form of the activation function from $x^2$ to the non-even functions $x + ax^2$ illustrated in Fig. 5 significantly affects the grokking behavior. Thus in Fig. 6, for P=27 and TRAIN_FRAC = 0.8, the number of epochs before grokking occurs decreases as the value of $a$ increases. Evidently, the odd, $x$, term delays the rise of the test accuracy and hence enhances the grokking behavior. Grokking also does not occur for odd activation functions $x^2\text{sign}(x)$ and $x^3$. In the latter case the test accuracy rises to a maximum value before decreasing, as evident in Fig. 7. In contrast, for a $|x^3|$ activation function, the test accuracy rises and then remains at a maximum value. Taken together, these results establish that a rapid onset of grokking requires that the activation function be even and that the onset of grokking can be delayed an arbitrary number of epochs by incorporating an odd, non-grokking component into the activation function.

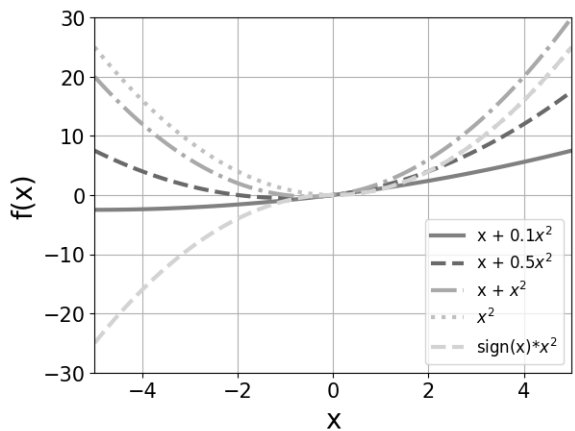

*Figure 5 Second-order polynomial activation functions.*

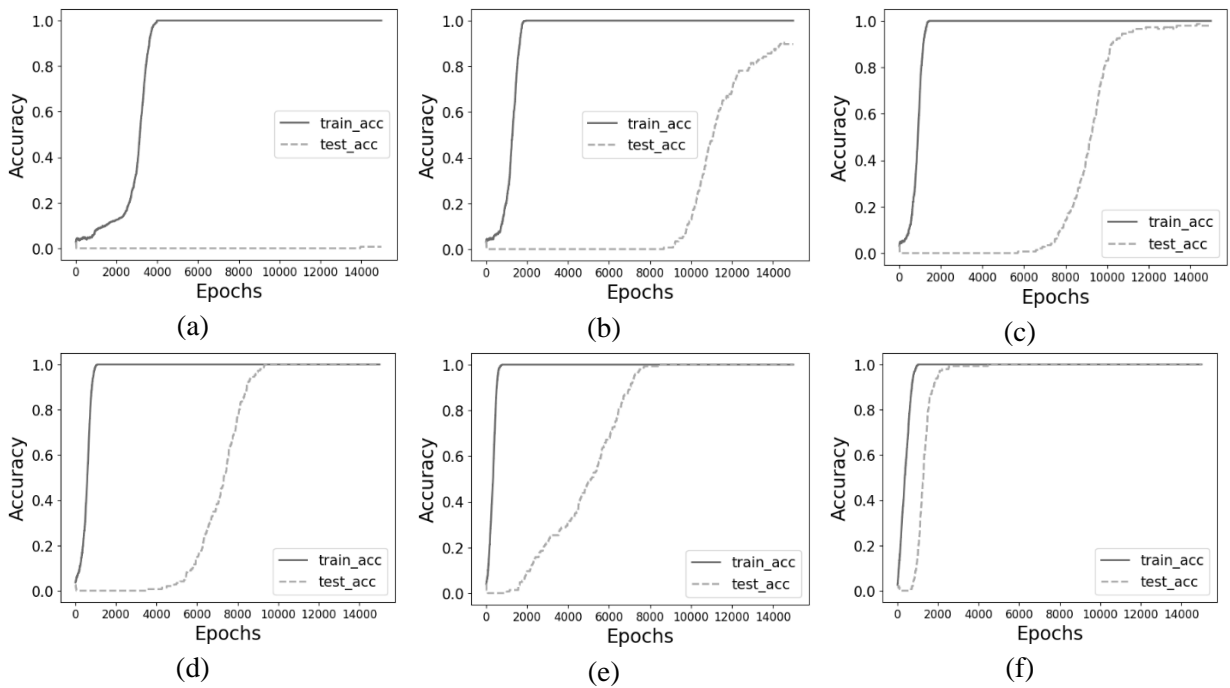

*Figure 6 The training/test accuracy for the activation functions (a) $x + 0.1x^2$ (b) $x + 0.5x^2$ (c) $x + x^2$ (d) $x + 2x^2$ (e) $x + 5x^2$ (f) $x^2$.*

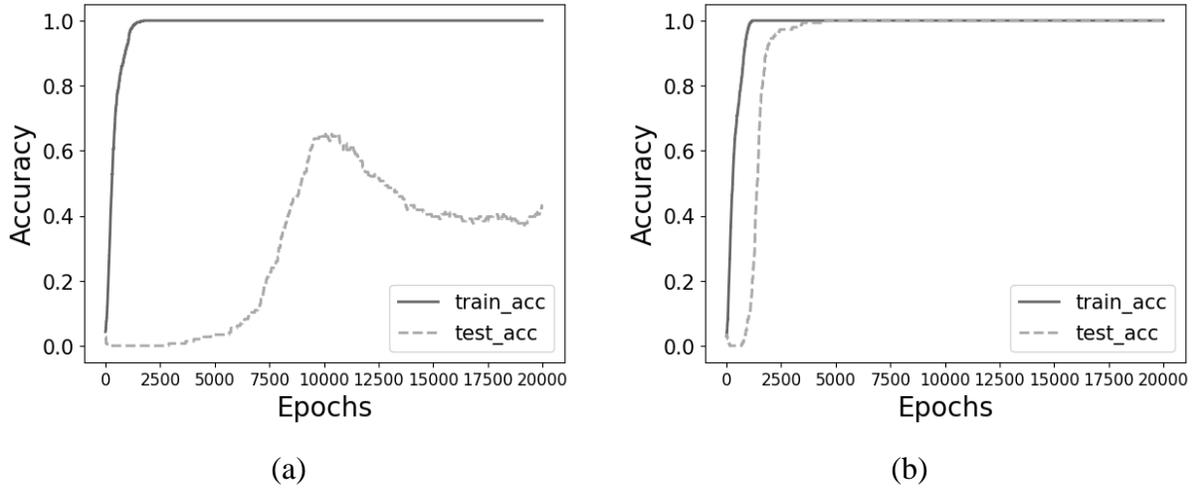

Figure 7 The training/test accuracy for the activation functions (a) $x^3$ (b) $|x^3|$.

The degree of nonlinearity of a network can also be adjusted by changing the number of neurons. Decreasing the number of neurons of the first layer from 256 to 64 for the case with $x^2$ activation function, reduces the slope of both the training and test accuracy curves as shown in Fig. 8 while the delay between the increases in test and training accuracies decreases, presumably since fewer neurons require optimization.

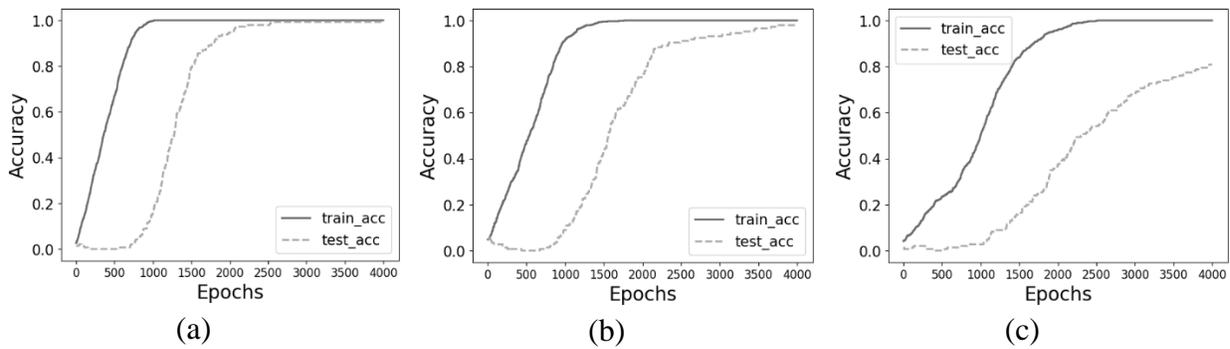

Figure 8 Training and test accuracies for (a) 256 (b) 128 and (c) 64 neurons.

## 3.3 PCA Projections

If an additional dense layer with neurons is added to the model and an activation function of the form $x + ax^2$, is employed for the first two layers, plotting the odd order PCA projections of the third layer weights against the even order projections as demonstrated by Power et al. (2022) [1], and Liu et al. (2022)[7] yields a circular pattern for some pairs of lower-order principal components. This structure is not present in a two-layer system, presumably because of insufficient nonlinearity. While grokking is observed for all $a$ values, a circular structure in PCA space only results when $a < 0.5$, as illustrated in Fig. 9, for $a = 0.25$, with P=53, TRAIN_FRAC = 0.5 and trained for 13k epochs in both cases. The corresponding structures for the higher-order PCA components are localized near the origin, as evident from Fig. 9.

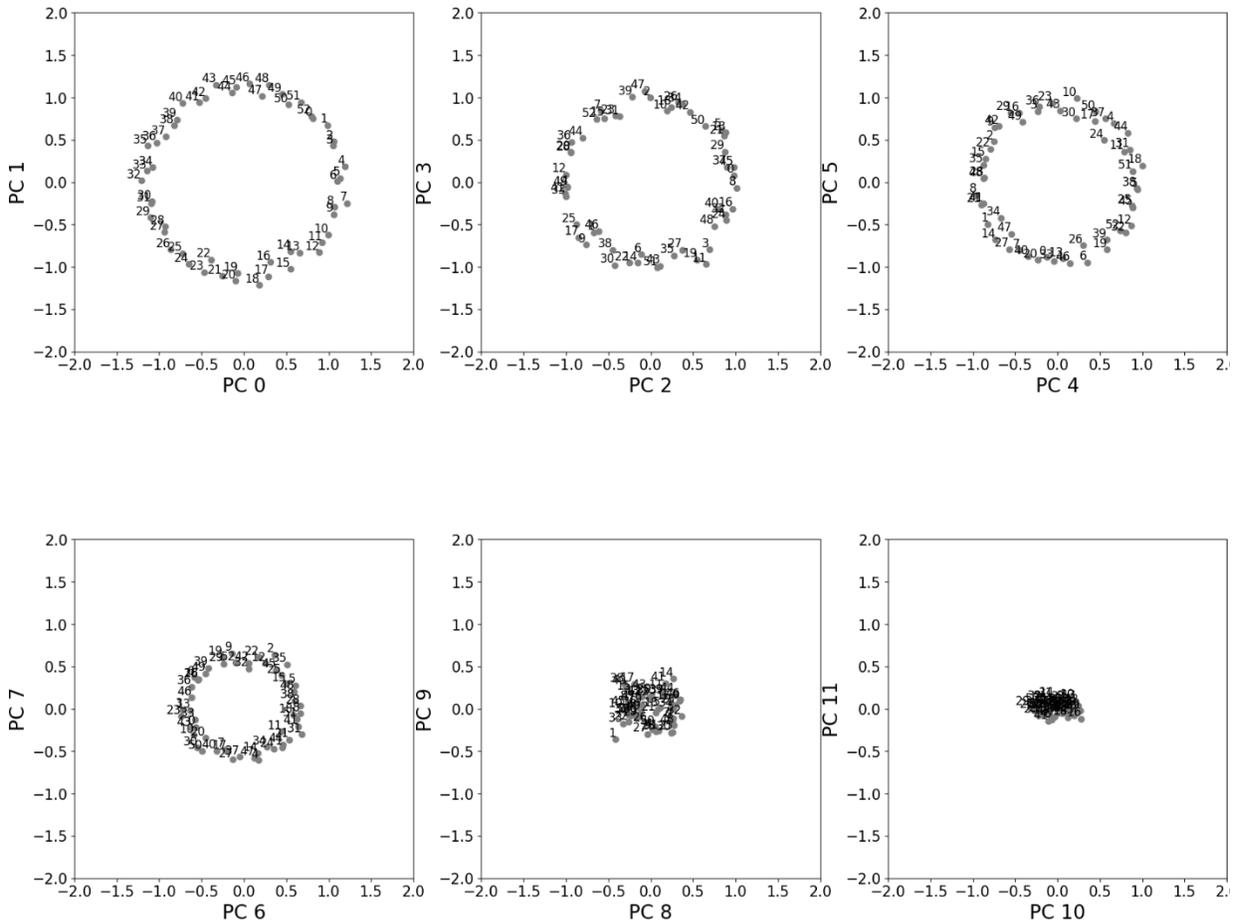

*Figure 9 Plots formed from successive PCA projections of the weights of the last layer after applying the activation function $x + 0.25x^2$ to the first and middle layers.*

*3.4 PCA Factoring*

Decreasing P to 20 reduces the size of the dataset and hence the effect of grokking. If the ratio of the number of samples in the training set to the total number of samples, TRAIN_FRAC = 0.7 the test accuracy decreases after reaching 1 so that TRAIN_FRAC = 0.8 is instead employed. While 64 neurons are then sufficient to produce grokking, successive groups of PCA projections of weights then do not display a circular structure. With 256 neurons in the two hidden layers, plotting the projections of the output patterns onto the zeroth and first and second and third lowest PCA components in Fig. 10 yields a circular pattern containing 5 clusters of 4 points while the pattern generated by the 4[th] and 5[th] lowest order PCA components contains 4 clusters with 5 points. Accordingly, the procedure has led to a factorization of 20 as $5 \times 4$.

The activation function $x + 0.1x^2$ in place of $x + 0.25x^2$ yields a different circular PCA space pattern generated by the projections on the 4[th] and 5[th] lowest order PCA components. Here the clusters consist of pairs of numbers that differ by 10 as illustrated in Fig. 11 yielding the factorization $20 = 2 \times 10$. Similarly, for P=21 the PCA space yields the 3 and 7 point clusters. In each of these cases, the patterns only appear after the program has executed a certain number of epochs beyond the occurrence of grokking.

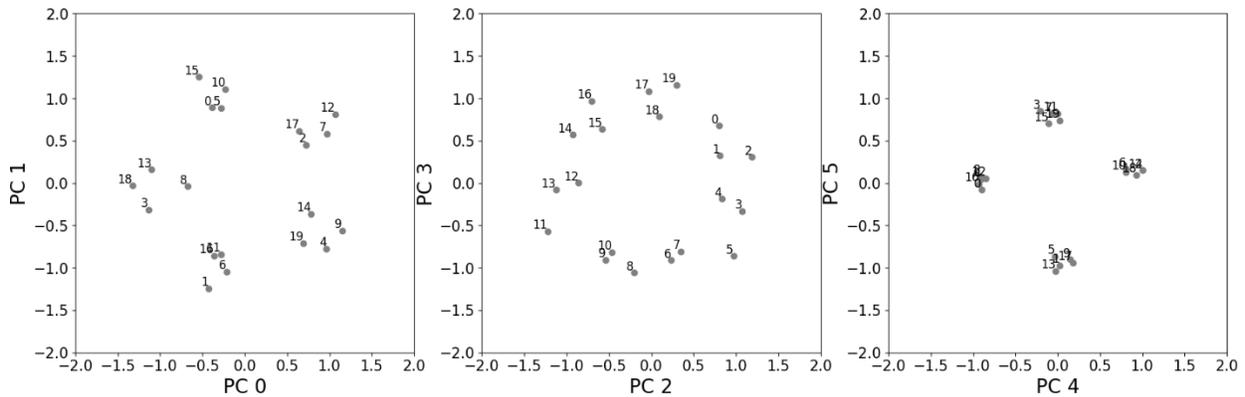

*Figure 10 Same as Fig.9 but with P=20 corresponding to the factoring $20 = 5 \times 4$.*

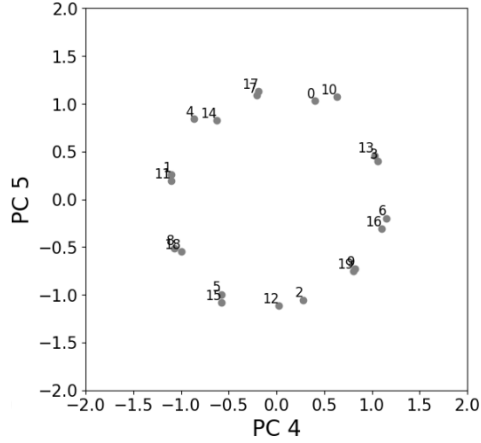

Figure 11 Same as Fig.10 but for the activation function $x + 0.1x^2$.

### 3.5 Data Dependence

Grokking in the modular addition problem could be explained by symmetries in the addition table or by the concatenated hot encoding of the two input values. One symmetry in the dataset is associated with the commutation relation $i + j = j + i$, and can be removed by considering only the values of $i + j$ with $j \geq i$, which, however, reduces the size of the dataset. In this case, for P = 53 and two network layers the maximum test accuracy is 0.85 after grokking as shown in Fig. 12 (b) for an $x^2$ activation function. Similar results are obtained for a $x + 2x^2$ activation function or for a 3 layer model.

If the data set is expanded to the same number of values that would have been generated if both $i + j$ and $j + i$ had been employed for P=53 by increasing P to 75, then for an $x^2$ activation function the test accuracy is substantially increased but does not reach unity as evident in Fig. 12(e). If the training set fraction is instead increased to 0.8 instead of 0.5, generalization occurs for P=53. The symmetry of the data set can be further reduced by, e.g. excluding the pairs with $j - i = 1, 2, 5$ and 8, which is however found not to affect the grokking unless the additional restriction that $i > j$ is also imposed. In the latter case, grokking is absent unless the training set fraction is set to 0.8, which then yields a maximum test accuracy of 0.984.

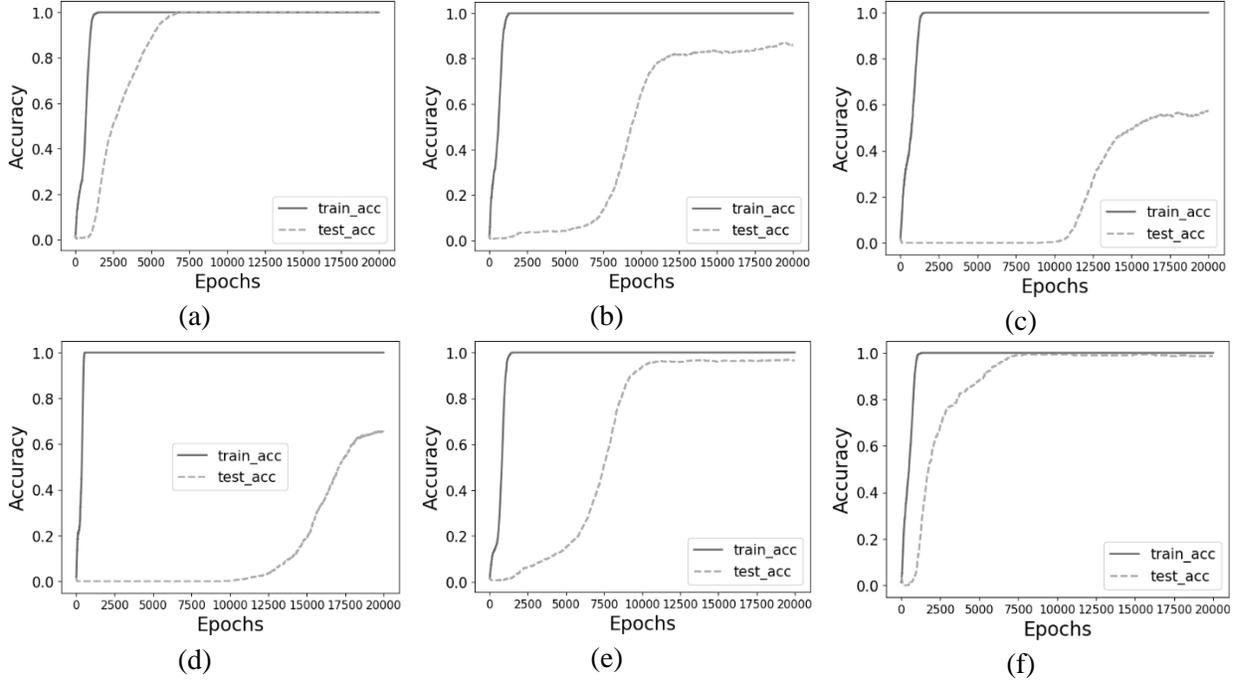

Figure 12 (a) Grokking for P = 53, TRAIN_FRAC =0.5 and an $x^2$ activation function (b) As in (a) but for $i + j$ with $i > j$ (c) as in (b) but for a $x + 2x^2$ activation function (d) as in (b) but for 3 NN layers (e) as in (b) but for P = 75 (f) as in (b) but with TRAIN_FRAC =0.8.

### 3.6 *Entropy Analysis*

The entropy of the weights of a NN layer, defined as $S = -\sum_i w_i log w_i$, where $w_i$ is the weight of each neuron in the layer, provides a metric for the extent of learning during training as the entropy quantifies the level of randomness in the weight distribution. Evenly distributed weights result in large entropy values indicating that the network is still exploring different configurations or is becoming unnecessarily complex leading to overfitting. For the simplest case of a $x^2$ activation function with P=53 and TRAIN_FRAC =0.5, Fig. 13 demonstrates that the entropy initially decreases while the training accuracy increases as the network learns relevant features. However, the entropy then increases further as the test accuracy starts to grok and only decreases again once the test accuracy reaches a maximum value. Hence the entropy is predictive of the additional learning that is required before the network can correctly distinguish the unseen test patterns. As further illustrated in Fig. 14, for an activation function of the form $x + x^2$, with P=53 and TRAIN_FRAC =0.8, after the initial rise in training accuracy, the decrease in entropy is less pronounced until it falls rapidly after the test accuracy reaches a maximum. If a third layer is employed, the entropy decreases to lower values than in the two layer network, presumably as a result of the increased nonlinearity. The initial fluctuations in the entropy during training are also dependent on the network nonlinearity, in agreement with Notsawo et al. (2023)[15], who predicted the number of epochs required to observe grokking from the loss landscape during early training stages.

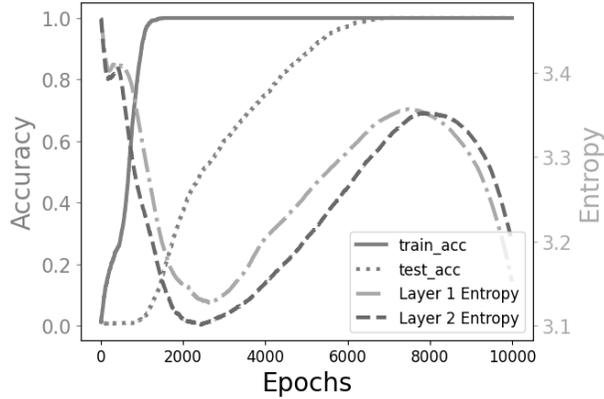

*Figure 13 The entropies of the first (dash dots) and second (dashed line) layer weights as a function of epoch number plotted together with the test (solid) and training (dots) accuracies as a function of epoch number for an $x^2$ activation function.*

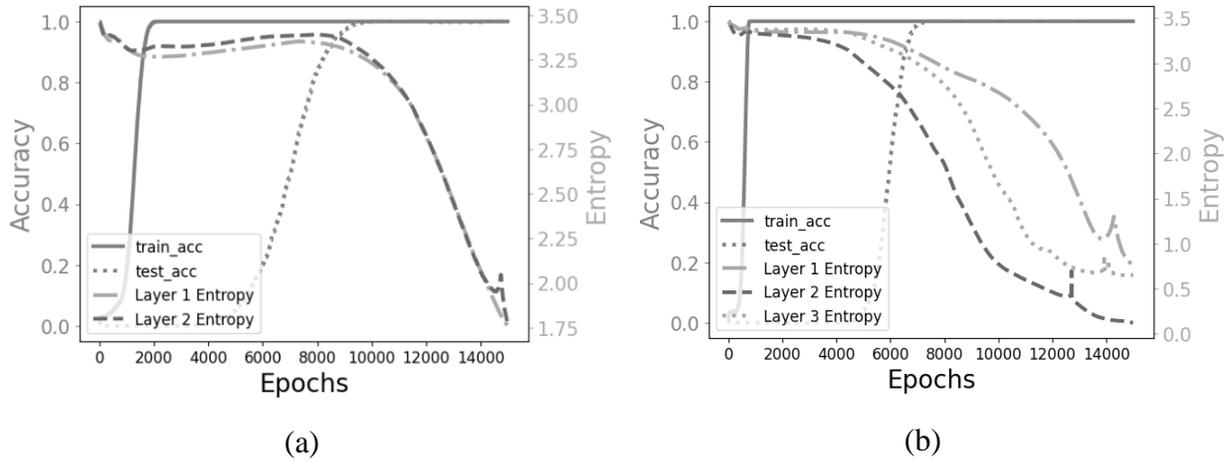

(a)                                     (b)

*Figure 14 As in the previous figure but for (a) an activation function of the form $x + x^2$ (b) an additional intermediate layer.*

## 4. Conclusion

This paper has examined the relationship between the properties of a simple neural network and its nonlinear structure in the context of modular addition. The nonlinearity was controlled through the form of the activation function, the number of layers and the number of neurons per layer. A procedure was identified based on these studies that enabled the number of epochs before the onset of grokking to be adjusted by incorporating a linear component into the activation function. In a similar manner, increasing the total nonlinearity by increasing the width or depth of the network was demonstrated to increase the delay between the rises in training and test accuracy. A mechanism for factoring non-prime numbers for networks with a sufficient number of layers and hence degree of nonlinearity was then proposed based on the PCA projections of the weights of the last layer. These results were then shown to apply even when the symmetries inherent in the dataset were partially eliminated if a larger fraction of the data is employed as training samples. Finally, metrics were advanced based on both the correlation between the weights of the connections to a neuron in the last layer and on the entropy of the neurons in each layer. These can be employed to quantify the learning process of the NN and its grokking behavior.

Further studies of these topics could further elucidate the relationship between the overall nonlinearity and the number of epochs required to achieve grokking as well as the predictive ability of the correlation and entropy metrics. A more systematic examination of the relationship between the correlation function of the neural network weights and the structures that arise in PCA space could potentially also be of practical significance. Finally, the degree to which the features of the highly simplified model examined in this paper are present in large scale networks that exhibit grokking and the associated similarities and differences between the two models should be examined in detail.

[3] Neel Nanda, Lawrence Chan, Tom Liberum, Jess Smith, and Jacob Steinhardt. Progress measures for grokking via mechanistic interpretability. arXiv preprint arXiv:2301.05217, 2023.

[4] Vimal Thilak, Etai Littwin, Shuangfei Zhai, Omid Saremi, Roni Paiss, and Joshua Susskind. The slingshot mechanism: An empirical study of adaptive optimizers and the grokking phenomenon. arXiv preprint arXiv:2206.04817, 2022.

[5] Vikrant Varma, Rohin Shah, Zachary Kenton, János Kramár, and Ramana Kumar. Explaining grokking through circuit efficiency. arXiv preprint arXiv:2309.02390, 2023.

[6] Ziming Liu, Eric J Michaud, and Max Tegmark. Omnigrok: Grokking beyond algorithmic data. arXiv preprint arXiv:2210.01117, 2022.

[7] Ziming Liu, Ouail Kitouni, Niklas Nolte, Eric J Michaud, Max Tegmark, and Mike Williams. Towards understanding grokking: An effective theory of representation learning. arXiv preprint arXiv:2205.10343, 2022.

[8] Andrey Gromov. Grokking modular arithmetic. arXiv preprint arXiv:2301.02679, 2023.

[9] Schaeffer, R., Miranda, B., and Koyejo, S. Are emergent abilities of large language models a mirage? In Thirty seventh Conference on Neural Information Processing Systems, 2023.

[10] Michaud, E. J., Liu, Z., Girit, U., and Tegmark, M. The quantization model of neural scaling. In Thirty-seventh Conference on Neural Information Processing Systems, 2023.

[11] Kumar, Tanishq, Blake Bordelon, Samuel J. Gershman, and Cengiz Pehlevan. "Grokking as the transition from lazy to rich training dynamics." *arXiv preprint arXiv:2310.06110* (2023).

[12] Loshchilov, I. and Hutter, F. Decoupled weight decay regularization. In 7th International Conference on Learning Representations, ICLR 2019, New Orleans, LA, USA, May 6-9, 2019.

[13] David Yevick, "Nonlinearity Enhanced Adaptive Activation Function," *arXiv preprint arXiv:2403.19896,* 2024.

[14] Boaz Barak, Benjamin L Edelman, Surbhi Goel, Sham Kakade, Eran Malach, and Cyril Zhang. Hidden progress in deep learning: Sgd learns parities near the computational limit. arXiv preprint arXiv:2207.08799, 2022.

[15] Pascal Jr. Tikeng Notsawo, Hattie Zhou, Mohammad Pezeshki, Irina Rish, and Guillaume Dumas. Predicting grokking long before it happens: A look into the loss landscape of models which grok, 2023.